\documentclass[letterpaper, 10 pt, conference]{ieeeconf}  

\IEEEoverridecommandlockouts                              

\usepackage{gensymb,soul}
\usepackage{graphics} 
\usepackage{subfigure}
\usepackage{booktabs}
\usepackage{epsfig} 
\usepackage{cite}
\usepackage{mathptmx} 
\usepackage{amsfonts,amssymb, amsmath}  
\usepackage[colorlinks,linkcolor=blue]{hyperref}

\usepackage{xcolor}
\usepackage{ifthen}
\usepackage{makecell}
\usepackage{algorithmicx}
\usepackage{algpseudocode}
\usepackage{threeparttable}
\usepackage[linesnumbered,ruled]{algorithm2e}

\DeclareMathAlphabet{\mathcal}{OMS}{cmsy}{m}{n}
\DeclareSymbolFont{largesymbols}{OMX}{cmex}{m}{n}

\usepackage{multirow}

\makeatletter
\let\NAT@parse\undefined
\makeatother

\title{\LARGE \bf
Robot Navigation in Unknown and Cluttered Workspace with Dynamical System Modulation in Starshaped Roadmap
}

\author{Kai Chen, Haichao Liu, Yulin Li, Jianghua Duan, Lei Zhu, and Jun Ma, \textit{Senior Member, IEEE}
\thanks{Kai Chen, Haichao Liu, and Lei Zhu are with the Robotics and Autonomous Systems Thrust, The Hong Kong University of Science and Technology (Guangzhou), Guangzhou 511453, China (e-mail: kchen916@connect.ust.hk; hliu369@connect.ust.hk; leizhu@ust.hk). }
\thanks{Jianghua Duan is with the Department of Mechanical and Aerospace Engineering, The Hong Kong University of Science and Technology, Hong Kong SAR, China (e-mail: jhduan@ust.hk).}
\thanks{Yulin Li and Jun Ma are with the Robotics and Autonomous Systems Thrust, The Hong Kong University of Science and Technology (Guangzhou), Guangzhou 511453, China, and also with the Division of Emerging Interdisciplinary Areas, The Hong Kong University of Science and Technology, Hong Kong SAR, China (e-mail: yline@connect.ust.hk; jun.ma@ust.hk). 
\textit{(Corresponding author: Jun Ma.)}}}

\begin{document}

\maketitle
\thispagestyle{empty}
\pagestyle{empty}

\begin{abstract}

Compared to conventional decomposition methods that use ellipses or polygons to represent free space, starshaped representation can better capture the natural distribution of sensor data, thereby exploiting a larger portion of traversable space. 
This paper introduces a novel motion planning and control framework for navigating robots in unknown and cluttered environments using a dynamically constructed starshaped roadmap. Our approach generates a starshaped representation of the surrounding free space from real-time sensor data using piece-wise polynomials. Additionally, an incremental roadmap maintaining the connectivity information is constructed, and a searching algorithm efficiently selects short-term goals on this roadmap. Importantly, this framework addresses dead-end situations with a graph updating mechanism.
To ensure safe and efficient movement within the starshaped roadmap, we propose a reactive controller based on Dynamic System Modulation (DSM). This controller facilitates smooth motion within starshaped regions and their intersections, avoiding conservative and short-sighted behaviors and allowing the system to handle intricate obstacle configurations in unknown and cluttered environments.
Comprehensive evaluations in both simulations and real-world experiments show that the proposed method achieves higher success rates and reduced travel times compared to other methods. It effectively manages intricate obstacle configurations, avoiding conservative and myopic behaviors. The source code will be released on website\footnote{Available at:~\href{https://github.com/kkkkkaiai/starshaped_roadmap}{github.com/kkkkkaiai/starshaped\_roadmap}}. 
\end{abstract}

\section{Introduction}

\begin{figure}[tbp]
    \centering
    \includegraphics[width=0.95\linewidth]{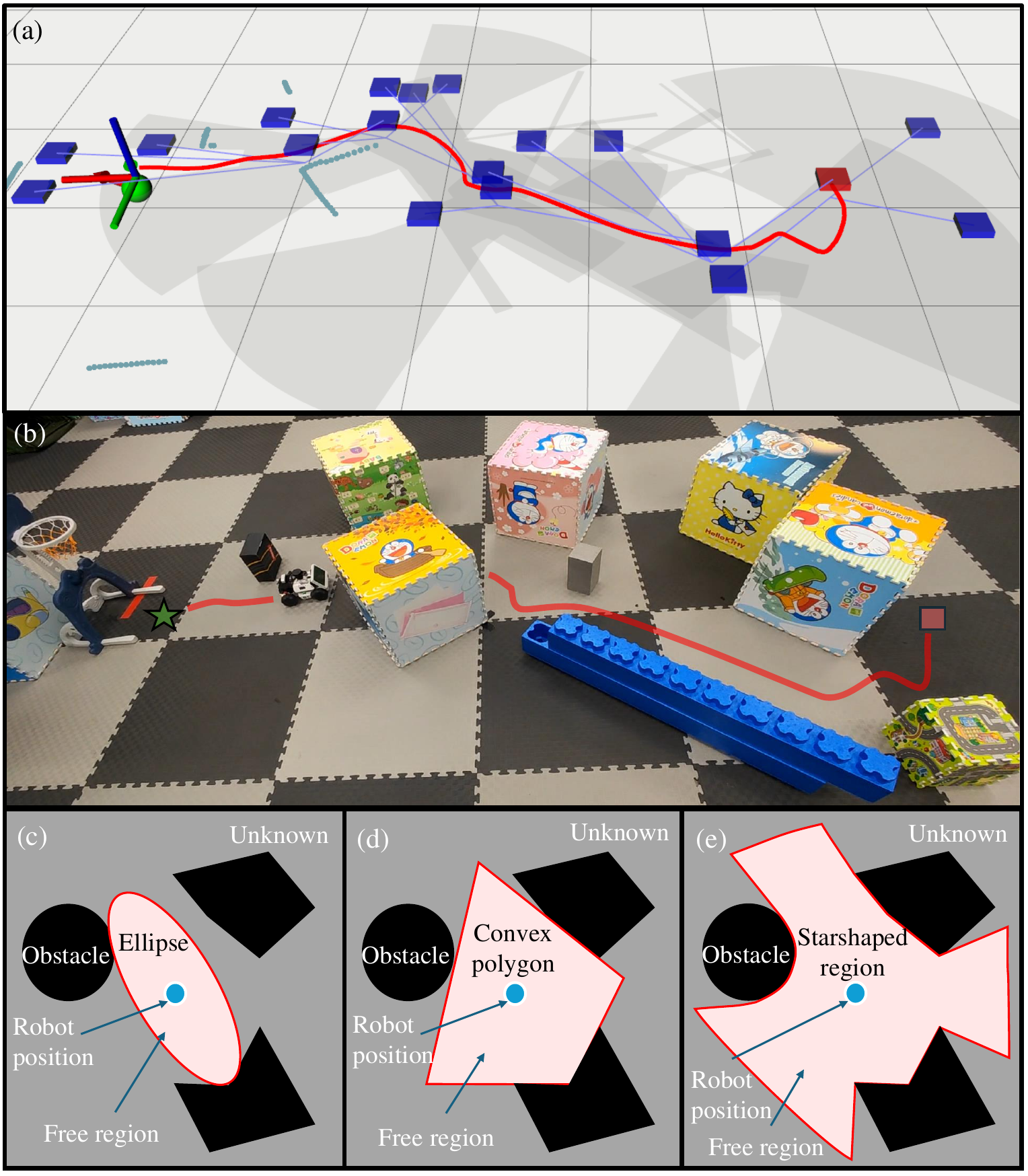}
    \caption{Highlight of the proposed navigation framework. \textbf{(a)} Our method, a dynamic starshaped roadmap is constructed in real-time to facilitate the proposed reactive controller based on DSM, enabling the robot to navigate smoothly towards its goal. \textbf{(b)} With our approach, the robot successfully navigates through unknown and cluttered environments, demonstrating its effectiveness in real-world scenarios. \textbf{(c)-(e)} Comparing with the representation of free regions by ellipses and convex polygons, our approach uses starshaped regions, covering a larger area and enabling better use of perception information for navigation.
    }
    \vspace{-19pt}
    \label{fig:illustration_real}
\end{figure}

The field of robot motion planning and control has made significant strides~\cite{guo2023recent}, with applications across diverse scenarios. However, navigating robots in partially or fully unknown environments, particularly those dense with obstacles, remains a substantial challenge. This study presents a framework specifically designed for rapid navigation in completely unknown and cluttered environments.
 
As shown in Fig.~\ref{fig:illustration_real}, our proposed approach generates starshaped regions to represent the surrounding free space directly from locally perceived point cloud data. By converting this data in polar coordinates, we facilitate fitting with piece-wise polynomial functions. Then, we dynamically construct a starshaped roadmap, where the frontier points of each region are identified and stored as potential extension points.
 
During the navigation task, a sequence of frontier points as coarse path points is searched on the roadmap. Short-term goals are then continually selected from these path points, allowing the robot to efficiently approach the target. Importantly, situations where the robot may get stuck due to a greedy search are effectively addressed with a roadmap updating mechanism. We also customize the dynamic System Modulation (DSM) method for our starshaped roadmap, enabling mobile robots to achieve smooth, efficient movement within overlapping starshaped free regions. The method is tested in a series of challenging navigation tasks, demonstrating that safe, fast, goal-oriented maneuvers are achieved with a high success rate.

\begin{figure*}[htbp]
    \centering
    \includegraphics[width=0.95\linewidth]{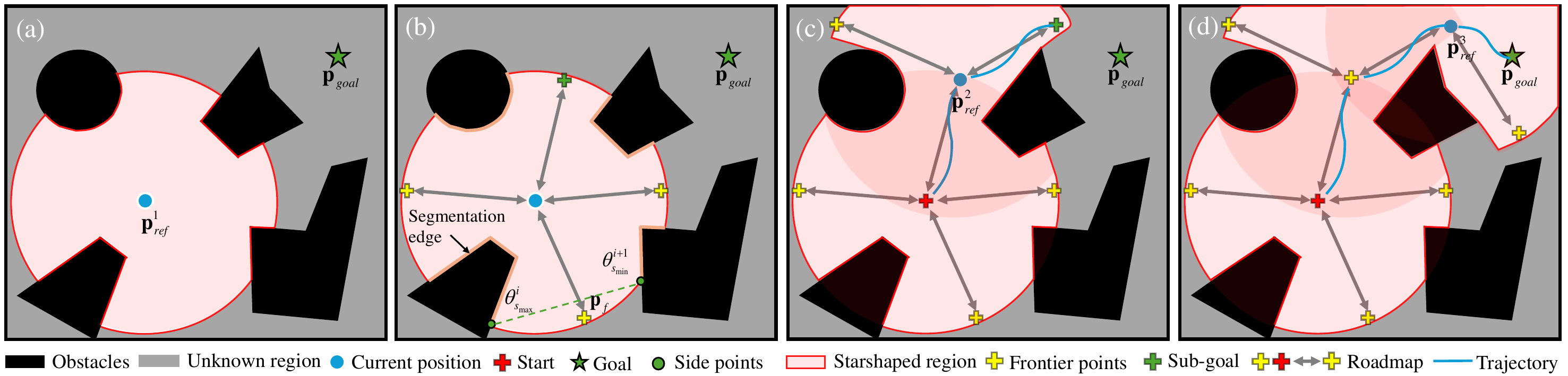}
    \caption{Illustration of the navigation process on our dynamically constructed starshaped roadmap. \textbf{(a)} The roadmap is initialized with the starshaped region at the start position. \textbf{(b)} Frontier points are generated on the roadmap. \textbf{(c)} The short-term goal is subsequently selected based on the search algorithm, and a new starshaped traversable region is extracted upon reaching the current short-term goal if it is extendable. \textbf{(d)} The navigation process is carried on with repeatedly calculated short-term goals till the target is reached.}
    \label{fig:roadmap_pipeline}
    \vspace{-18pt}
\end{figure*}

Our contributions are summarized as follows:
\begin{itemize}
    \item We propose a novel method for representing traversable space in unknown and cluttered environments by constructing starshaped regions from locally perceived point cloud data. This approach maximizes the use of available free space, avoids conservative behavior, and dynamically generates a roadmap with extendable regions.
    \item We introduce a reactive controller based on DSM for our starshaped roadmaps, which is operated at a high frequency of up to 200 Hz to facilitate smooth and efficient robot movement.
    \item Extensive experiments, conducted both in simulation and on real-world robots, validate the effectiveness of the proposed method, demonstrating its robustness and applicability in practical scenarios. 
\end{itemize}

\section{Related Works}

The concept of Safe Corridors (SC)~\cite{faster_mit, park2020online, ding2019safe} has been extensively used in recent robot planning frameworks. These frameworks generate a sequence of overlapping convex shapes to efficiently approximate collision-free space, allowing effective enforcement of collision avoidance constraints due to the convexity of the free regions. For instance, a robust bi-level optimization process proposed in~\cite{ciris} iteratively finds the free polytope and its minimum enclosing ellipsoid separating the robot from obstacles while maximizing the ellipsoid's volume. This method effectively covers a large portion of the free space but requires high computation time, limiting its use in real-time applications.
Subsequent methods based on point clouds~\cite{decomp} and voxel grids~\cite{toumieh2022voxel} allow the real-time generation of SC with large volumes. However, these convex shapes may not fully utilize the environment information from sensors, leading to conservative and myopic behavior, especially in narrow and cluttered spaces.
In contrast, starshaped regions~\cite{starshaped_rd, creatingstar} naturally cover a larger portion of free space within the perception range. Yet, efficiently parameterizing and extracting starshaped free region representations in real-time and generating safe and effective motion within these regions remains a significant challenge.

In partially or fully unknown environments, local feedback controllers are favored by researchers for their ability to quickly react to unexpected changes using real-time onboard sensor data. These approaches often rely on distance field representations, such as the Euclidean Signed Distance Field (ESDF) \cite{chmop} and Artificial Potential Field (APF)\cite{apf}. The encoded obstacle information can be used directly \cite{safeAPF} or integrated into an optimization framework \cite{esdf_opt} to adjust the robot's motion, promoting obstacle-avoiding or goal-reaching behaviors. However, potential field based methods suffer from local minima and add computational overhead due to preprocessing of generating the fields from sensor data.
On the other hand, DSM has gained popularity for its timely obstacle avoidance capabilities by adjusting initial system dynamics ~\cite{huber2022avoiding, khansari2012dynamical, saveriano2014distance, dsmwithmpc, foa}. DSM has been extended to handle overlapping convex obstacles and starshaped obstacles \cite{huber2022avoiding}. However, these methods require prior knowledge of obstacle geometries to calculate the modulation matrix and have not been implemented for navigation in fully unknown and cluttered environments, struggling with complex obstacle layouts.
Alternatively, a reactive control law with rigorous safety guarantees has been developed to navigate robots in locally generated polytopic free regions~\cite{geometry}. However, this method requires a semi-algebraic expression of the free region, which is not trivial and is difficult to generalize to starshaped regions.

\section{Method}
\subsection{Problem Formulation}
The workspace of the environment is defined as $\mathcal{G}\subset \mathbb R^n$ and it contains obstacles $\mathcal{O}_i$, with $i =1,2,\cdots,M$. 
The free space of the environment is denoted as:
\begin{equation}
    \mathcal{W}=\mathcal{G}\backslash \bigcup_{i=1}^M\mathcal{O}_i\subset \mathbb R^n.
\end{equation}
Our objective is to identify the free space based on locally perceived point clouds. As described in Section \ref{sec:roadmap}, these identified regions, along with detected frontier points, are then used to construct a roadmap. This roadmap is subsequently integrated with control methods to enable a disk-shaped robot to navigate to its target while avoiding obstacles in the environment.

\subsection{{Starshaped Dynamic Roadmap Generation}}\label{sec:roadmap}
In this work, our roadmap serves two key purposes: it identifies potential paths for robot navigation and delineates the traversable areas within the space. The proposed roadmap, called starshaped roadmap, consists of nodes and edges. Each node stores a position and the traversable area generated from that position as the origin, representing the range within which the robot can move without collision. The edges define the direct connections between nodes, representing feasible transitions within the environment.

\subsubsection{Representation of the Traversable Regions}\label{star:param}
To effectively represent the traversable area, we employ the starshaped set to represent it. Our approach covers a larger area when expanding from the same point compared with the previous methods, and an intuitive visual comparison is shown in Fig.~\ref{fig:illustration_real}(c)-\ref{fig:illustration_real}(e). Before introducing our method, it is necessary to provide the definition of the starshaped set.

A set \( \mathcal{S} \subseteq \mathcal{W} \) is starshaped with respect to a point \( \mathbf{p}_{ref} \in \mathcal{S} \), for every point \( \mathbf{p} \in \mathcal{S} \), the line segment connecting \( \mathbf{p}_{ref} \) and \( \mathbf{p} \) lies entirely within \( \mathcal{S} \). Formally, this can be expressed as:
\begin{align*}
\mathcal{S} \text{ is starshaped} \iff &\exists \, \mathbf{p}_{ref} \in \mathcal{S} \text{ such that } 
    \, \forall \, \mathbf{p} \in \mathcal{S}, \\
    & (1 - \lambda) \mathbf{p}_{ref} + \lambda \mathbf{p} \in \mathcal{S} \text{ for all } \lambda \in [0, 1].
\end{align*}
Here, $\mathbf{p}_{ref}$ is referred to as the center of the starshaped set $\mathcal{S}$. 

A precise representation of the boundary of the starshaped region is essential. After the boundary is established, the distance from any position in the space to the boundary can be determined. In this study, the robot obtains point clouds from a 2D LiDAR sensor,  which is used to fit the boundary of the starshaped region. The position of the LiDAR sensor is used as the center point $\mathbf{p}_{ref}$, and the line segments connecting the scanned points satisfy the conditions of a starshaped set.  

Subsequently, we employ piece-wise polynomials to describe the boundary. We first transform the point cloud $\mathcal{L}_{\mathbf{p}_{ref}}$ from the Cartesian coordinate system into the polar coordinate system, with the origin of the point cloud at $\mathbf{p}_{ref}$. One key advantage of using the polar coordinate system is that any fitting inaccuracies only affect the precision of the boundary and do not alter its compliance with the starshaped definition. Each point $\mathbf{p}_l=(p_{x,l},p_{y,l})\in\mathcal{L}$ to the polar coordinates $(\theta,d)$, where $\theta$ represents the angular direction and and $d$ denotes the radial distance from $\mathbf{p}_{ref}$. For any position within the region, excluding $\mathbf{p}_{ref}$, the transformation is given by the following formulation:
\begin{equation}\label{eq:cartesianToPolar}
\left[\begin{array}{l}
\theta \vspace{0mm} \\
d
\end{array}\right]
=
\left[\begin{array}{c}
\arctan(p_{y,l}-p_{y,{ref}}, p_{x,l}-p_{x,{ref}})  \vspace{1.5mm} \\
\sqrt{(p_{y,l}-p_{y,{ref}})^2+(p_{x,l}-p_{x,{ref}})^2} 
\end{array}\right].
\end{equation}

Each segment of the piece-wise polynomials is represented by a $N$th-order polynomial, defined as follows:
\begin{equation}\label{equ:polinomial}
\phi(\theta)=a_0+a_1\theta+a_2\theta^2+\cdots+a_N\theta^N = \sum_{i=0}^{N} a_{i} \theta^i.
\end{equation}
The complete piece-wise function, composed of these polynomials, is defined as:
\begin{equation}\label{eq:piece-wisefunc}
\Phi(\theta)=
\begin{cases}
\sum_{i=0}^{N} a_{i,1} \theta^i, & -\pi \le \theta \le \theta_1, \\
\sum_{i=0}^{N} a_{i,2} \theta^i, & \theta_1 < \theta \le \theta_2, \\
\vdots & \\
\sum_{i=0}^{N} a_{i,k} \theta^i, & \theta_{k-1} < \theta \le \pi.
\end{cases}
\end{equation}

Consequently, we use the $\{\theta, d\}$ pairs from (\ref{eq:cartesianToPolar}) and apply the least squares method to estimate the coefficients of the polynomials.

\subsubsection{Frontier Points Generation}\label{sec:fp}
Similar to some exploration algorithms, our newly generated nodes originate from frontier points that are unexplored but have potential for further expansion. In this part, our object is to identify the central positions between obstacles on the starshaped boundary, which will serve as the frontier points.

Our proposed method, illustrated in Fig.~\ref{fig:roadmap_pipeline}(b), consists of the following steps: First, for the point cloud $\mathcal{L}$, we remove points with the maximum range of the LiDAR, as these points do not correspond to detected obstacles. Next, we apply the DBSCAN algorithm~\cite{dbscan} to cluster the remaining points and remove outliers. From each cluster, we extract two points, referred to as \texttt{side points}. These represent the outermost points in polar coordinates within the cluster, characterized by their angular values $\theta_s\in [-\pi, \pi]$. 
Finally, we connect adjacent side points from neighboring clusters and compute their geometric midpoint to determine the frontier points, which represent the boundary between the explored and unexplored regions. Mathematically, each frontier point is defined by its angular value:
\begin{equation}
\theta_f =\frac{\theta_{s_{max}}^{i}+\theta_{s_{min}}^{i+1}}{2},
\end{equation} 
where $\theta_{s_{max}}^{i}$ is the maximum angular value from cluster $i$, and $\theta_{s_{min}}^{i+1}$ is the minimum angular value from the adjacent cluster $i+1$. The location of the frontier point can then be computed using the inverse transformation of ($\ref{eq:cartesianToPolar}$). The set of all frontier points is denoted as $\mathcal{F_S}$, which forms the frontier of the starshaped set $\mathcal{S}$. 

According to the above definition, the dynamic roadmap $\mathcal{R(N,E)}$ is constructed as follows: The position of the first node $n_0 \in \mathcal{N}$ is set as the robot's initial state, and a starshaped region is created using the current perception data. The positions of newly generated frontier points, forming part of nodes $n_i \in \mathcal{N}$, are then added to the roadmap, and the edges $e_i \in \mathcal{E}$ between $n_0$ and $n_i$ are recorded. Each time the robot reaches an unvisited node, the process is repeated.

\subsubsection{Roadmap Update}
The objective of updating the roadmap is to address the challenge of managing detected frontier points that cannot be further expanded. Frontier points in the roadmap are characterized by two attributes: \texttt{extendable} and \texttt{stuck}. Initially, each point is designated as \texttt{extendable}, but this status may change as the robot approaches the point. The evaluation process is as follows: when the robot arrives at a frontier point, it applies the method outlined in Section \ref{sec:fp} to generate new frontier points. If new frontier points cannot be generated, or if the newly generated points fall within the starshaped region of another node, the frontier point is reclassified as \texttt{stuck}.

\begin{algorithm}[t]
\caption{Robot Navigation with DSM in Starshaped Roadmap}
\label{alg:DSM}
\SetKwFunction{MyFunction}{GenRoadmap}
\SetKwProg{Fn}{Function}{:}{}
\KwIn{$\mathbf{p}$, $\mathbf{p}_{goal}$, $\mathcal{L}$ and $\mathcal{R}$}
\KwOut{$\mathbf{u}$ and $\mathcal{R}$}
\DontPrintSemicolon
\Fn{\MyFunction{$\mathbf{p}, \mathcal{L}, \mathcal{R}$}}{
    $\mathcal{P} \gets$ {Fit piece-wise polynomial with $\mathcal{L}$}; \;
    $\mathcal{F} \gets$ Identify frontier points with $\mathcal{L}$; \;
    $\mathcal{N} \gets$ $\mathbf{p}$ and $\mathcal{F}$; \;
    $\mathcal{E} \gets$ Constructing edge set in $\mathcal{N}$; \;
    $\mathcal{R}^* \gets$ Build the roadmap with $\mathcal{N}$ and $\mathcal{E}$; \;
    $\mathcal{R} \gets \mathcal{R} \cup \mathcal{R}^*$; \;
    \KwRet{$\mathcal{R}$}
}
Initialize $\mathcal{R}$ as $\emptyset$ and $p_{sg}$ does not exist; \;
\While {$\mathbf{p}$ {not reach} $\mathbf{p}_{goal}$}{
\If{$\mathcal{R}$ is $\emptyset$ or $\mathbf{p}=\mathbf{p}_{sg}$ or \texttt{update}}{
    $\mathcal{R} \gets$ \MyFunction{$\mathbf{p},\mathcal{B},\mathcal{R}$}
}

$\mathbf{p}_{sg} \gets$ Search in the roadmap $\mathcal{R}$; \;
    \texttt{stuck} $\gets$ Check the feasibility of $\mathbf{p}_{sg}$; \;
    \If{\texttt{stuck}}{
        Delete $\mathbf{p}_{sg}$ in $\mathcal{R}$; \;
        \texttt{replan}, \texttt{update} $\gets$ $True$
    }
    \If{$\mathbf{p}=\mathbf{p}_{goal}$ or \texttt{replan}}{
        Continue; \;
    } 
$\Tilde{\mathbf{p}}^* \gets$ Get modulated guiding vector with $\mathbf{p}_{sg}$ using (\ref{eq:sum_weighted}); \;
$\Tilde{\mathbf{p}}^* \gets$ Get combined guiding vector using (\ref{equ:combine}); \;
$\mathbf{u} \gets$ Get control input with $\Tilde{\mathbf{p}}^*$ using a low-level controller;
}
\end{algorithm}

\subsection{Navigation in Starshaped Roadmap}\label{subsec:dsm}

In recent work~\cite{huber2022avoiding}, reactive control with obstacle avoidance is achieved within starshaped regions. However, their method requires prior knowledge of obstacle shapes or the formulation of predefined starshaped regions, making it impractical for fully unknown environments. To address this limitation, we propose a reactive controller used in our proposed starshaped roadmap based on DSM, allowing its use in unknown environments.

\subsubsection{Robot Model}
The robot is modeled as a disk-shaped differential drive agent with a radius of \(r\). Its state is represented by its position \(\mathbf{p} = (p_x, p_y) \in \mathbb{R}^2\), while the control input is defined as \(\mathbf{u} = (v, w) \in \mathbb{R}^2\), where \(v\) denotes the linear velocity and \(w\) the angular velocity. During the navigation process, a desired guiding vector $\Tilde{\mathbf{p}}$ is always generated, guiding the robot from its current state to the target position. 

\subsubsection{Short-Term Goal Selection}
Given an arbitrary starting position \(\mathbf{p}\), typically the robot's current location, and a goal position \(\mathbf{p}_{\text{goal}}\), the robot navigates step-by-step towards the goal by searching through a sequence of nodes from the frontier points in the roadmap. At each step, the first node in the sequence, excluding the start node, is treated as a short-term goal $\mathbf{p}_{sg}$, guiding the robot to move within a starshaped region. We denote this sequence of nodes as \( \mathbf{p}_{f,i} \in \mathcal{F} \), where \( i \in \{1, 2, \cdots, M\} \) and \( M \) represents the number of selected nodes during the search process. To compute this sequence, a general graph search method is employed, and the corresponding mathematical formulation is presented below:
\begin{equation}\label{eq:optim_selection}
\begin{array}{cl}
\underset{{\mathbf{p}_{f,1}, \mathbf{p}_{f,2}, \cdots, \mathbf{p}_{f,M}}}{\min} &\sum_{i=1}^{M} \|\mathbf{p}_{f,i} - \mathbf{p}_{f,i-1}\| \\
\text{s.t.} & \mathbf{p}_{f,1} = \mathbf{p},\ \mathbf{p}_{f,M} = \mathbf{p}_{\text{goal}}, \\
& (\mathbf{p}_{f,i}, \mathbf{p}_{f,i+1}) \in \mathcal{E}, \quad \forall i \in \{1, 2, \cdots, M-1\},
\end{array}
\end{equation}
where the third constraint ensures that \( \mathbf{p}_{f,i} \) and \( \mathbf{p}_{f,i+1} \) exist an edge in the roadmap \( \mathcal{E} \). It is important to note that nodes labeled as \texttt{stuck} are excluded from this search process.

\subsubsection{Motion Control in Starshaped Roadmap}
In this subsection, we develop a DSM-based feedback reactive controller to navigate towards the selected short-term goal $\mathbf{p}_{sg}$.
Our goal is to modulate the guiding vector \( \tilde{\mathbf{p}} \) to ensure safe navigation on the constructed starshape roadmap. 

At the robot position $\mathbf{p}$, the current guiding vector is given by:
\begin{equation}
\tilde{\mathbf{p}} = 
\mathbf{p}_{sg} - \mathbf{p},
\end{equation}
and the modulation of \(\tilde{\mathbf{p}} \) is defined as:
\begin{equation}\label{eq:modulate}
    \tilde{\mathbf{p}}^* = \mathbf{M}(\mathbf{p}) \tilde{\mathbf{p}},
\end{equation}
where \( \mathbf{M}(\mathbf{p})=\mathbf{E}(\mathbf{p}) \mathbf{D}(\mathbf{p}) \mathbf{E}(\mathbf{p})^{-1} \) is the modulation matrix that adjusts the direction of motion, thereby enabling obstacle avoidance by preventing the robot from reaching the detected boundaries of the starshaped regions. 
The matrix \( \mathbf{E}(\mathbf{p}) \) contains eigenvectors derived from the starshaped representation, and \( \mathbf{D}(\mathbf{p}) \) is a diagonal matrix containing the eigenvalues that encode how the velocity should be scaled or altered in each principal direction. These matrices are given as:
\begin{equation}\label{equ:dsm_matrix}
\begin{split}
    \mathbf{E}(\mathbf{p})&=[\mathbf{r}(\mathbf{p})^\top,
    {\mathbf{e}_1(\mathbf{p})^\top, \cdots,\mathbf{e}_{n-1}(\mathbf{p})^\top}]^\top\in \mathbb R^{n\times n},\\
    \mathbf{D}(\mathbf{p})&=\operatorname{diag}[\lambda_r(\mathbf{p}), \underbrace{\lambda_{e}(\mathbf{p}),\cdots, \lambda_{e}(\mathbf{p})}_{n-1}] \in \mathbb R^{n\times n},
\end{split}
\end{equation}
where $\mathbf{r}(\mathbf{p})=\mathbf{p}_{ref}-\mathbf{p}$ is the eigenvector pointing toward the reference point, associated with the scaling factor \( \lambda_r(\mathbf{p}) \),  which governs the reduction of velocity in the critical direction towards the boundary. The remaining eigenvectors \( \mathbf{e}_1(\mathbf{p}), \dots, \mathbf{e}_{n-1}(\mathbf{p}) \) are orthogonal to \( \mathbf{r}(\mathbf{p}) \) and are associated with the scaling factor \( \lambda_e(\mathbf{p}) \), which allows greater freedom of motion in safer directions.

According to the analytical expression of the parameterization of starshaped sets in Section \ref{star:param}, the vector in the normal direction can be computed as:  
\begin{equation}\label{eq:elementOfBasisMat}
    \mathbf{e}_i = [\cos(\Bar{\theta}_i), -\sin(\Bar{\theta}_i)]^\top, 
\end{equation}
where
\begin{align}
\Bar{\theta} &= \frac{
\Dot{\Phi}(\theta)\sin(\theta) + \Phi(\theta)\cos(\theta)
}{
\Dot{\Phi}(\theta)\cos(\theta) - \Phi(\theta)\sin(\theta)
},  \
\theta = \arctan\left( \frac{p_y - p_{ref,y}}{{p}_x - p_{ref,x}} \right).
\end{align}
Here, $\Dot{\Phi}(\theta)$ represents the first-order derivative of (\ref{eq:piece-wisefunc}) with respect to $\theta$. The scaling factors in (\ref{equ:dsm_matrix}) are calculated by:
\begin{equation}\label{eq:elementOfEigenvalueMat}
\begin{array}{l}
\lambda_r(\mathbf{p}) = 1 - \frac{1}{\Gamma(\mathbf{p})}, \
\lambda_e(\mathbf{p}) = 1 + \frac{1}{\Gamma(\mathbf{p})},
\end{array}
\end{equation}
where \( \Gamma(\mathbf{p}) = \left(\frac{\Phi(\mathbf{p})}{\|\mathbf{p}-\mathbf{p}_r\|}\right)^\sigma \) for \( \mathbf{p} \neq \mathbf{p}_r \), and \( \infty \) otherwise. The parameter \( \sigma \) is predefined.

\begin{figure}[tbp]
    \centering
    \includegraphics[width=1.0\linewidth]{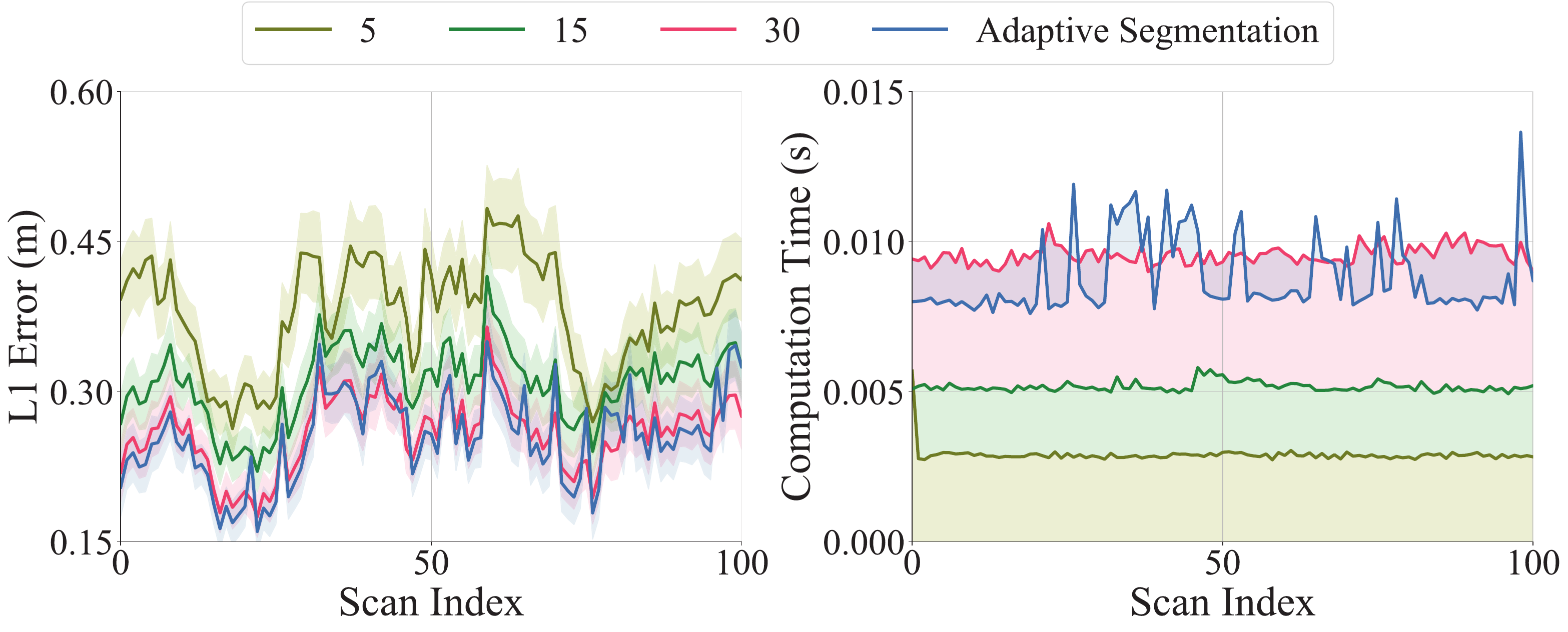}
    \caption{Ablation study of the proposed adaptive segmentation method for piecewise polynomials. Compared to different fixed segment numbers, our method achieves higher accuracy while offering a trade-off in computation time.}
    \label{fig:ada_seg}
\end{figure}

As illustrated in Fig.~\ref{fig:roadmap_pipeline}\,(d), when the robot navigates the environment, it may encounter overlapping areas of multiple starshaped regions. To guide the robot through these overlaps, we assign weights to each overlapping starshaped set. The weight for each set is determined by its individual \(\Gamma_k(p)\) value, relative to the total weight within the $M$ overlaps:

\begin{equation}
w_k(\boldsymbol{\mathbf{p}})=\frac{\max \left\{\Gamma_k(\boldsymbol{\mathbf{p}}), 1\right\}}{\sum_{k=1}^{M} \max \left\{\Gamma_k(\boldsymbol{\mathbf{p}}), 1\right\}}.
\end{equation}
By integrating the weights of the starshaped sets in the overlapping area, the robot's final control input $\Tilde{p}^*$ can be computed as follows:
\begin{equation}\label{eq:sum_weighted}
    \Tilde{\mathbf{p}}^* = \sum_{k=0}^{M}w_k\mathbf{M}_k(\mathbf{p})\Tilde{\mathbf{p}}.
\end{equation}


Considering the radius of the robot's shape, the guiding vector is further refined using the nearest point \( \mathbf{p}_n \) from the received point cloud \(\mathcal{L}_\mathbf{p}\):
\begin{equation}
    {\mathbf{p}}^* = (1-\alpha)\Tilde{\mathbf{p}}^*+\alpha\Tilde{\mathbf{p}}_n^*, \quad \alpha = \min\left(\frac{\rho}{||\mathbf{p}-\mathbf{p}_n||-r}, 1\right),
    \label{equ:combine}
\end{equation}
where $\rho$ is a predefined parameter. A low-level controller follows this guiding vector to achieve effective navigation. 

The overall algorithm of the proposed method is outlined in Algorithm~\ref{alg:DSM}.

\section{Experiments and Results}

This section compares our proposed method with other DSM-related approaches, including DSM\cite{huber2022avoiding}, FOA\cite{foa}, and FOA with our proposed roadmap (FOA with RM). The evaluation criteria include travel time, path length, computation time, and success rate. Performance is assessed through both simulation and real-world experiments, using a differential drive mobile robot tested in cluttered environments. Both experiments are conducted using Python 3.10 and ROS on an Intel i7-12700H processor. It is important to note that, once the modulated guiding vector is calculated according to~(\ref{equ:combine}), a PID controller is employed to control the robot.

\begin{figure}[tbp]
\centering
\subfigure[A forest-like cluttered scenario with cylindrical obstacles.]{
	\label{Fig.sub.1}
	\includegraphics[width=0.47\linewidth]{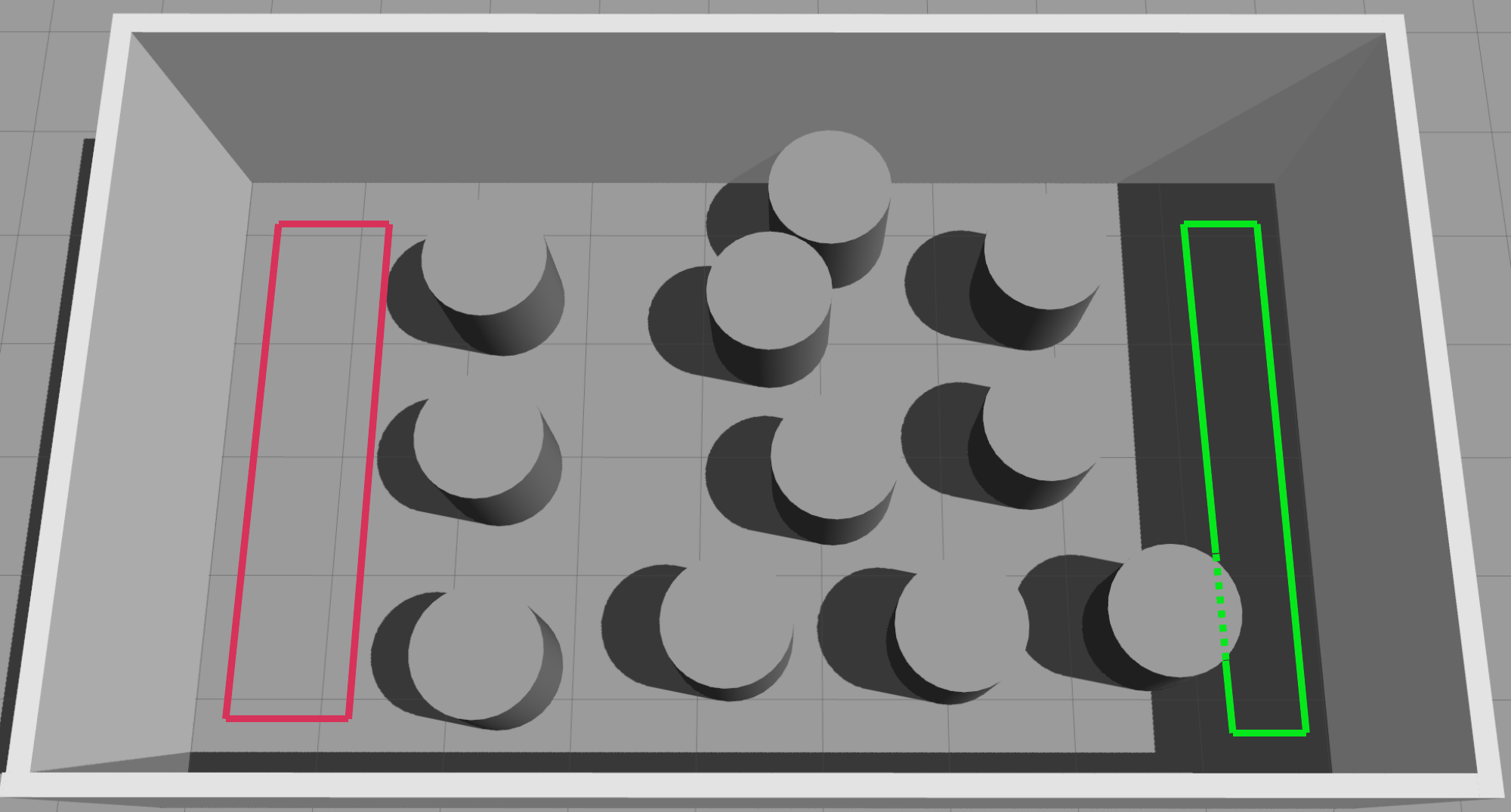}}
\subfigure[A challenging maze-like scenario with spherical, cylindrical, and concave obstacles.]{
	\label{Fig.sub.2}
	\includegraphics[width=0.47\linewidth]{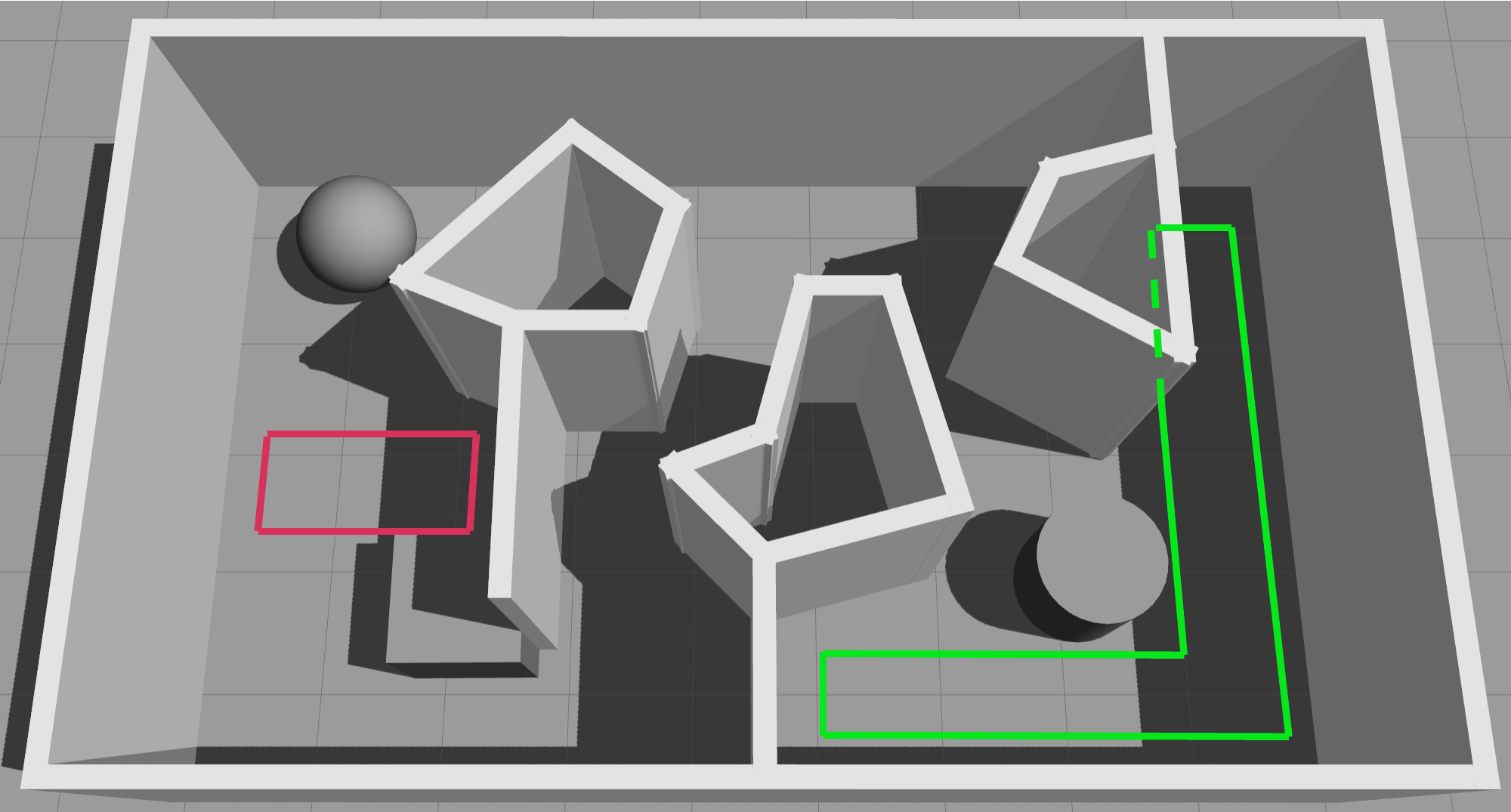}}

\caption{In the simulation experiments, the start and target positions will be randomly generated in the corresponding red and green regions.}
\label{fig:scenarios}
\end{figure}

\begin{table}[t]
\caption{Comparison of the navigation performance}
\centering
\begin{threeparttable}
\resizebox{0.95\linewidth}{!}{
\begin{tabular}{@{}ccccc@{}}
\toprule
\toprule
\textbf{Scenario} & \multicolumn{1}{c}{\textbf{Method}} & \textbf{TTR Avg. (\%)} & \textbf{PLR Avg. (\%)} & \textbf{SR (\%)} \\ 
\midrule
\multirow{4}{*}{\makecell{Forest}}   & \multicolumn{1}{c}{DSM\cite{huber2022avoiding} }   &  \textbf{89.8}  &  102.14    &   73.3                \\
    & \multicolumn{1}{c}{FOA\cite{foa}}        &       99.6       &    109.9    &    20.7 \\
    & \multicolumn{1}{c}{FOA with RM}   &       137.0    &    108.6    &        40       \\
    & \multicolumn{1}{c}{\textbf{Ours}}       &       100      &   \textbf{100}     &        \textbf{90}       \\ \midrule

\multirow{4}{*}{\makecell[c]{Maze}} & DSM\cite{huber2022avoiding}    &   --      &      --     &       --        \\
\multicolumn{1}{l}{} & FOA\cite{foa}        &     --       &     --      &      --      \\ 
\multicolumn{1}{l}{} & FOA with RM   &   115.8    &     \textbf{94.9}          &       80        \\
\multicolumn{1}{l}{} & \textbf{Ours}      &   \textbf{100}    &     100    &       \textbf{83.3}         \\ 
\bottomrule
\bottomrule
\end{tabular}
}
\end{threeparttable}\label{tab:simu_result}
\end{table}

To improve the accuracy of boundary fitting for starshaped regions in Section~\ref{star:param}, we utilize the rate of change in radial distance to adaptively determine the polynomial segments, balancing accuracy and computation time. In this case, $N=3$.
The error performance, in comparison with those obtained using a fixed number of segments, are presented in Fig.~\ref{fig:ada_seg}.

\subsection{Simulation Evaluation}

To evaluate the effectiveness and versatility of our method, we performe simulations using the Gazebo simulator in two representative scenarios. In each scenario, we generate different initial robot poses and target positions, repeating the experiment 30 times per scenario, as illustrated in Fig.~\ref{fig:scenarios}. 
To quantify the results, we define two performance metrics: the Travel Time Ratio (TTR) and the Path Length Ratio (PLR). TTR and PLR represent the travel time and path length ratio, where the benchmark method is in the numerator compared to our approach. As shown in Table~\ref{tab:simu_result}, the average values of TTR, PLR and Success Rate (SR) are recorded for all experiments in which the robot reaches the goal successfully.

\begin{figure*}[!htbp]
\centering
\subfigure[In the forest simulation, all methods except for FOA enable the robot to successfully navigate to the destination.]{
	\label{Fig:sub.1}
	\includegraphics[width=0.99\textwidth]{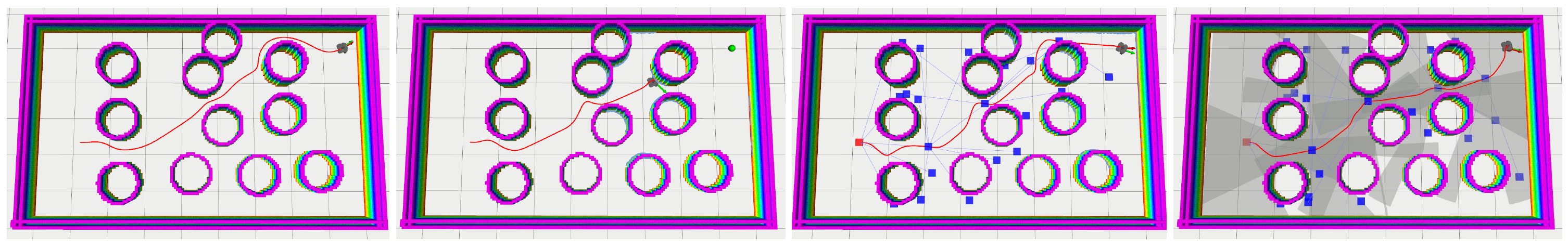}}
\subfigure[In the maze environment, both DSM and FOA do not allow the robot to reach the goal position.]{
	\label{Fig:sub.2}
	\includegraphics[width=0.99\textwidth]{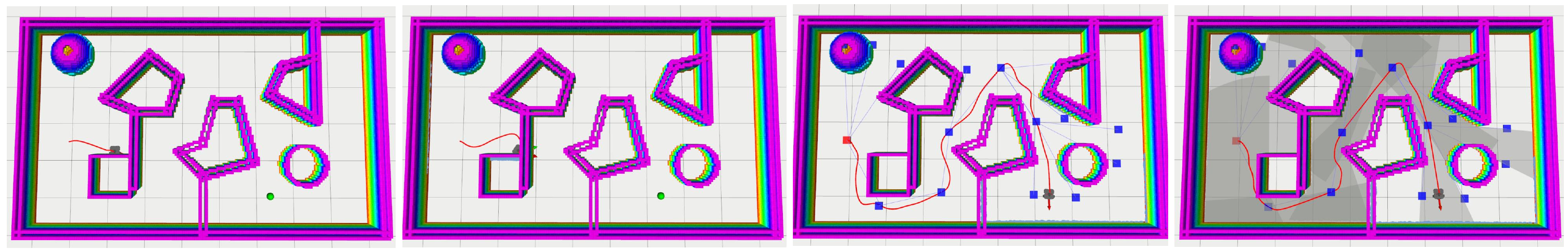}}
\caption{The simulation results, with methods from left to right: DSM, FOA, FOA combined with RM, and our approach, demonstrate that our method effectively handles both environments.}
\label{fig:senarios_result}
\vspace{-8pt}
\end{figure*}

\subsubsection{Forest Scenario} 

The scenario includes cylinders with a radius of 0.5\,m, and the varying spacing between the cylinders can create situations where the robot risks getting trapped into local minima. Notably, in this setup, DSM requires precise knowledge of the obstacles' shapes and positions. In contrast, for other methods, the environment remains entirely unknown. Our method achieves the highest success rate compared to the alternatives. Although DSM leverages prior knowledge of the environment for navigation, closely spaced cylinders can lead to local minima. However, the roadmap generated by our approach significantly enhances the robot's ability to escape from local minima. However, the FOA method frequently encounters saddle points, which prevent the robot from avoiding obstacles in time. Even with the assistance of roadmap guidance, the success rate of FOA shows only limited improvement.

\subsubsection{Maze Scenario}
 
In the maze scenario, the presence of walls and irregularly shaped objects creates a constrained environment with narrow passages, dead-ends, and potential bottlenecks. For DSM, the complexity of expressing intricate obstacles through functions makes it impossible to compute the modulation matrix in such environments. Meanwhile, FOA struggles because it relies solely on sensor data from the current frame, limiting its ability to handle the maze effectively. However, when combined with the starshaped roadmap introduced in this work, FOA's performance improves significantly. Despite this enhancement, our method still outperforms, providing faster navigation and higher success. 

\subsubsection{Computation Time} The average computing time for controlling the robot is $3.98\pm 0.96$ ms, and $4.37\pm 1.16$ ms in the forest and maze scenarios, respectively. The computational efficiency allows us to set a high control frequency, more than 200 Hz, for the robot to perform rapid reactions. The fitting process for the starshaped representation takes approximately $8\pm1.3$ ms, as shown in Fig.~\ref{fig:ada_seg}, and it only needs to be executed when expanding the roadmap and therefore has minimal impact on the overall computational efficiency of the proposed algorithm.

\begin{figure}[tbp]
    \centering
    \includegraphics[width=0.99\linewidth]{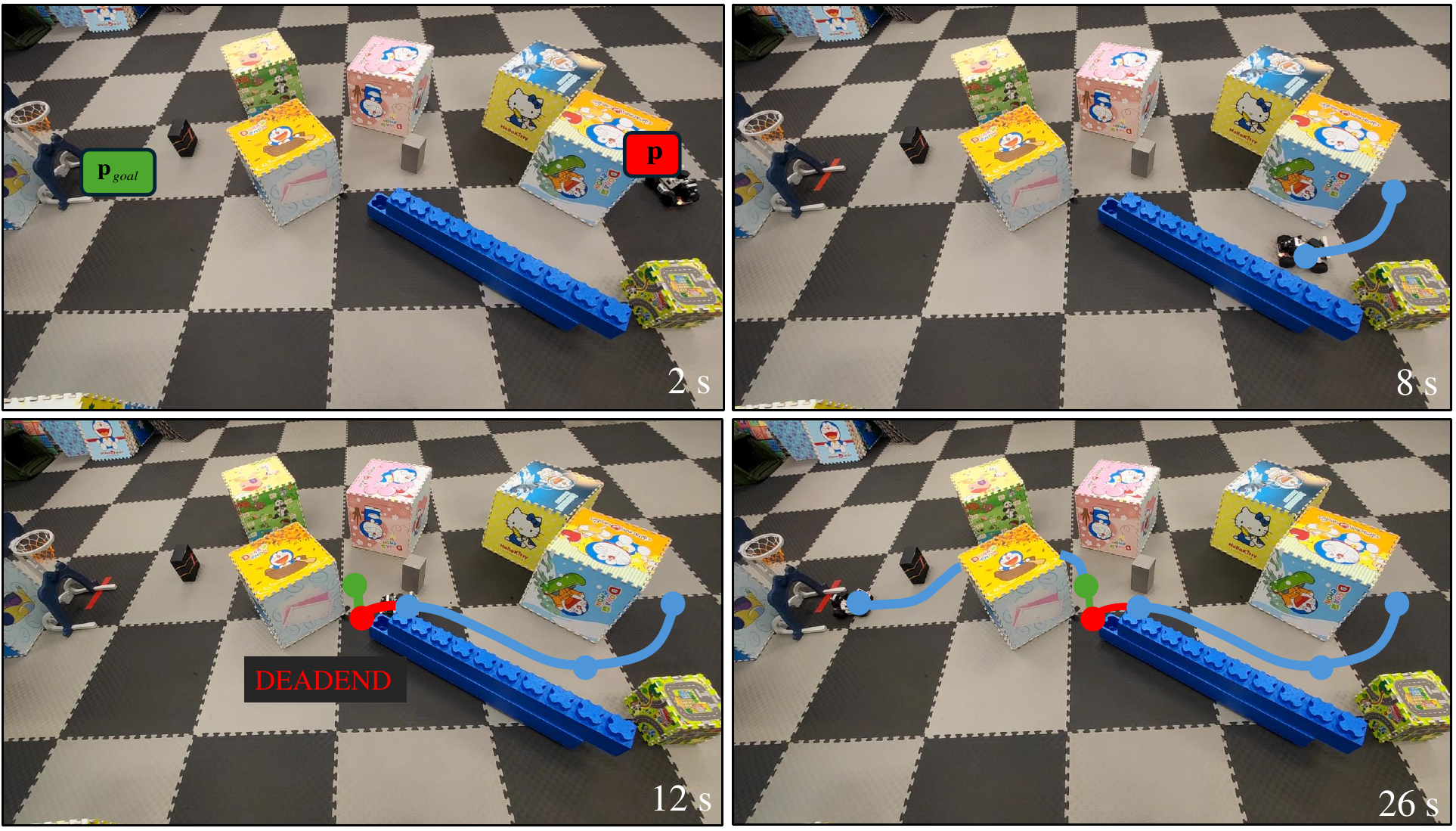}
    \caption{A real-world experiment conducted on a differential drive mobile robot, which successfully navigates the environment until encountering a dead-end scenario.}
    \label{fig:real_result}
\vspace{-8pt}
\end{figure}

\subsection{Real-World Experiment}

To validate our method in real-world conditions, we conduct experiments using a differential drive mobile robot equipped with a 2D LiDAR for environmental perception. The robot navigates through a \(6\,\text{m} \times 8\,\text{m}\) area populated with various obstacles. In the experiment, depicted in Fig.~\ref{fig:real_result}, the robot successfully maneuvers around randomly placed obstacles, reaching its destination. 
It takes an average of 10 ms to construct starshaped regions from the sensor data incrementally. The robot maintains an average speed of 0.4\,m/s, with control inputs calculated in 3\,ms on average.
The robot gets stuck around 12\,s, at which point the roadmap is updated immediately to identify a new short-term goal, allowing the robot to bypass the obstacle.


\section{Conclusion}

In this paper, we propose a novel motion planning and control framework for mobile robots that navigate in cluttered and unknown environments. Our proposed starshaped roadmap effectively utilizes starshaped regions derived from real-time point cloud data to represent free spaces. Meanwhile, our roadmap construction maintains connectivity information with frontier points, guiding the robot efficiently through a heuristic exploration algorithm. Then, we propose an effective reactive controller based on DSM to ensure smooth and safe movements within the starshaped regions for disk-shaped robots. Comparative experiments with state-of-the-art methods demonstrate that our approach yields better success rates and broader applicability. For future work, we aim to further develop the starshaped representation method to handle dynamic environments and explore more robust control strategies, such as model predictive control, to enhance stability and performance in the navigation task.

\bibliographystyle{IEEEtran}
\bibliography{mybibs}

\end{document}